\documentclass[manuscript]{acmart}
\AtBeginDocument{%
  \providecommand\BibTeX{{%
    \normalfont B\kern-0.5em{\scshape i\kern-0.25em b}\kern-0.8em\TeX}}}

\copyrightyear{2022}
\acmYear{2022}
\setcopyright{rightsretained}
\acmConference[FAccT '22]{2022 ACM Conference on Fairness, Accountability, and Transparency}{June 21--24, 2022}{Seoul, Republic of Korea}
\acmBooktitle{2022 ACM Conference on Fairness, Accountability, and Transparency (FAccT '22), June 21--24, 2022, Seoul, Republic of Korea}\acmDOI{10.1145/3531146.3533101}
\acmISBN{978-1-4503-9352-2/22/06}

\newcommand\group[1]{\texttt{#1}}
\newcommand\suggest[1]{\textbf{#1}}
\newcommand{\olga}[1]{{}}
\definecolor{amethyst}{rgb}{0.6, 0.4, 0.8}

\usepackage{array}
\usepackage{algorithm2e}
\usepackage{bbm}
\usepackage{graphicx}
\usepackage{wrapfig}
\usepackage{tabularx, booktabs}
\usepackage{multirow}
\newcommand{\smallsec}[1]{\vspace{.2cm} \noindent {\bf #1:}}

\begin{document}

\title{Towards Intersectionality in Machine Learning: Including More Identities, Handling Underrepresentation, and Performing Evaluation}
\renewcommand{\shorttitle}{Towards Intersectionality in Machine Learning}



\author{Angelina Wang}
\affiliation{%
  \institution{Princeton University}
  \country{USA}
}

\author{Vikram V. Ramaswamy}
\affiliation{%
  \institution{Princeton University}
  \country{USA}
}

\author{Olga Russakovsky}
\affiliation{%
  \institution{Princeton University}
  \country{USA}
}







\begin{abstract}
Research in machine learning fairness has historically considered a single binary demographic attribute; however, the reality is of course far more complicated. In this work, we grapple with questions that arise along three stages of the machine learning pipeline when incorporating intersectionality as multiple demographic attributes: (1) which demographic attributes to include as dataset labels, (2) how to handle the progressively smaller size of subgroups during model training, and (3) how to move beyond existing evaluation metrics when benchmarking model fairness for more subgroups. For each question, we provide thorough empirical evaluation on tabular datasets derived from the US Census, and present constructive recommendations for the machine learning community. First, we advocate for supplementing domain knowledge with empirical validation when choosing which demographic attribute labels to train on, while always evaluating on the full set of demographic attributes. Second, we warn against using data imbalance techniques without considering their normative implications and suggest an alternative using the structure in the data. Third, we introduce new evaluation metrics which are more appropriate for the intersectional setting. Overall, we provide substantive suggestions on three necessary (albeit not sufficient!) considerations when incorporating intersectionality into machine learning.

\end{abstract}



\maketitle

\section{Introduction}
As machine learning is being adopted in an increasing number of applications, there is a growing awareness and concern that people of different demographic groups may be treated unfairly~\cite{mitchell2018history}. Measuring and mitigating these effects often require assigning individuals to demographic groups, and this is frequently done along one axis of identity at a time, e.g., gender or race~\cite{friedler2019comparative}. However, when drawing boundaries and selecting demographic groups, it is important to recognize the intersectional harms that result from interacting systems of oppression.
\citet{crenshaw1989intersectionality} first coined the term ``intersectionality'' by showing that Black women experience discrimination beyond being either Black or women. 
Intersectionality broadly refers to how different identities along different axes interact to produce unique forms of discrimination and societal effects~\cite{collins2020intersectionality, carastathis2008privilege, crenshaw1989intersectionality}.\footnote{There are complex nuances to this conceptualization that are out of scope of this work~\cite{nash2008rethinking}.} 
There is a long history of considering intersectional harms in fields outside of computer science~\cite{combahee1977statement, king1988jeopardy, hooks1981woman, spelman1988inessential, collins1990empowerment, mccall2005complexity, nash2008rethinking, shields2008gender, collins2020intersectionality, makkonen2002fore, grillo1995antiessentialism}, and an urgent need to do so in machine learning fairness as well.
For example, \citet{kearns2018empirical} perform experiments that show the algorithmic harms intersectional subgroups may experience due to heterogeneity within a particular demographic group, e.g., \group{Female}. In other words, although a classifier may be fair with respect to gender, as well as race, it can be unfair with respect to the intersection of the groups, missing that, for example, \group{Black} \group{Female} and \group{White} \group{Female} may differ in substantial and meaningful ways~\cite{kay2021equality}.

In this work, we focus on the algorithmic effects of discrimination against demographic subgroups (rather than individuals). Specifically, we conduct empirical studies of five fairness algorithms~\cite{jiang2020reweight,kearns2018gerry,agarwal2018reductions,foulds2018intersectionality,yang2020overlap} across a suite of five tabular datasets derived from the US Census with target variables like income and travel time to work~\cite{ding2021retiringadult}. 
We do so under the framework of the canonical machine learning fairness setting: supervised binary classification of a target label of social importance, which balances accuracy and one mathematical notion of fairness among a finite set of discretely defined demographic groups, which may result from a conjunction of identities.\footnote{We acknowledge that the group identity delineations themselves are unstable and fraught with problems of operationalization~\cite{mccall2005complexity, hirschman2000race}.}

We echo the calls of prior work to consider multiple axes of identities~\cite{kearns2018empirical, foulds2018intersectionality}, but in this work, focus on the next steps \textit{after} someone has decided to consider intersectionality in their machine learning pipeline. In doing so, three core challenges emerge. First, in the dataset stage, we need to select which identity labels to consider. This is difficult because considering too many would be computationally intractable but considering too few may miss intersectional harms. Second, in the model training stage, we need to consider how to technically handle the progressively smaller number of individuals in each group that will result from adding additional identities and axes. Finally, in the evaluation stage, we need to decide how we will perform fairness evaluation as the number of groups increases. There are seemingly straightforward ways to address each of these three questions. For example, one might consider as many axes of identity as they have access to in the data; handle smaller groups by drawing from machine learning techniques for imbalanced data such as generating synthetic examples of underrepresented groups~\cite{ramaswamy2021latent, sattigeri2019gan, sharmanska2020contrastive}; and evaluate on more subgroups by generalizing existing fairness definitions, such as equal opportunity or demographic parity, through extrapolation~\cite{foulds2018intersectionality, kearns2018gerry}. However, by treating intersectionality as simply an extension of the binary group setting to a multi-group one, these straightforward approaches fail to critically engage with the substantive differences that intersectionality brings.

Our contributions in this work are in meaningfully engaging with these three problems that arise along different stages of the machine learning pipeline: dataset selection, model training, and model evaluation. These come after the decision to consider intersectionality, and our concrete suggestions are as follows:

\begin{enumerate}
    \item \textbf{Selecting which identities to include (Sec.~\ref{sec:heterogeneity}):}
    due to the tenuous nature of operationalizing demographic categories we will need to supplement domain knowledge with empirical results in order to understand which identities to include in model training.
    This applies not only to multiple axes, but also individual axes. 
    For example, when considering the racial group \group{Asian} \group{Pacific} \group{Islander}, there are many potential granularities of identities to include, such as breaking the group up into its constituent ones of \group{Hmong}, \group{Cambodian}, etc.
    We show that, in a way that is hard to know a priori, different algorithms benefit from training on different levels of granularity. 
    However, evaluation should generally be performed on as many demographic groups as are known.\footnote{There are concerns regarding noisy measurements of small groups that are out of scope for our work; we refer the reader to \citet{foulds2019measure}.} 

    \item \textbf{Handling progressively smaller groups (Sec.~\ref{sec:dataset}): } the more identities we consider, the smaller each group is likely to be. Normative concerns unrelated to the technical efficacy of data imbalance techniques can be enough to constrain or even disqualify their use; for example, harmful historical parallels connected to generating synthetic facial images can raise concerns. 
    We suggest a new path, hypothesizing that structure within intersectional data can be carefully exploited in very specific circumstances, such as by learning about statistical patterns in an underrepresented \group{Black} \group{Female} group from groups it might share characteristics with, like \group{Black} \group{Male}. 
    \item \textbf{Evaluating a large number of groups (Sec.~\ref{sec:evaluation}): } commonly used pairwise comparisons for fairness evaluation can obscure important information when extrapolated and applied to a greater number of subgroups. This precipitates a call for additional kinds of evaluation that measure considerations such as the reification of existing hierarchies amongst subgroups. For the algorithms and datasets we consider,
    we demonstrate that the ranking amongst subgroups for positive label base rates of the dataset is highly correlated with the rankings of true positive rates of the model predictions, even when training with fairness constraints.
\end{enumerate}

These considerations are not unique to intersectionality, as they are liable to arise in any multi-attribute setting, but considering intersectionality sharply precipitates their importance. We also note that despite the language we employ, we do not suggest that fairness can be treated as a purely algorithmic problem that neglects the sociotechnical frame~\cite{green2018myth, birhane2021injustice, green2020realism, selbst2019sociotechnical}. Like the limitation noted by recent work~\cite{ding2021retiringadult}, our contributions are limited to the realm of intersectional \textit{algorithmic fairness}, and not data-driven insights into societal intersectionality. Intersectionality is frequently considered through qualitative rather than quantitative approaches~\cite{atewologun2018practice} because of the flattening effect the latter has in treating groups as a monolith, so to an extent, quantitative studies will always be limited in this aspect.


\section{Related Work}
\label{sec:related_work}
The canonical machine learning fairness paradigm frequently assumes binary attributes along a single axis~\cite{friedler2019comparative}. For example, for the many algorithms that only work in this contrived setup, IBM's AI Fairness 360 tool~\cite{Bellamy18aif360} formulates the binary attribute as \group{White} and \group{Non-White}, a trend sometimes shared by the social sciences that may conceptualize of social categories as dichotomous, e.g., class as middle-class and poor, gender as men and women, and sexuality as heterosexual and homosexual~\cite{mattias2014transgender}. 
To get a high-level look at how prevalent the problem of not considering intersectionality is, we look at a set of 26 popular machine learning fairness algorithm papers.\footnote{We use Semantic Scholar to keyword search for ``fair'', ``fairness'', and ``bias'' from 9 conferences: NeurIPS, ICML, ICLR, FAccT, CVPR, ECCV, ICCV, ACL, EMNLP. We retained all papers with 75 or more citations, and of these 58 papers, further narrowed down to the 26 that proposed fairness algorithms.} Of these 26, only 16 can operate in a setting beyond binary attributes, and of those, only 7 report empirical results on multiple axes of identity.

Algorithmic fairness methods have begun to consider intersectional attributes beyond just one axis of identity~\cite{tan2019word, cabrera2019fairvis, steed2021unsupervised, kirk2021gpt}. \citet{kearns2018gerry} and \citet{yang2020overlap} offer learning methods for intersectional fairness, but weigh the fairness of each group by their frequency and thus downweigh underrepresented groups, which arguably should be the focus of intersectional fairness. \citet{hebertjohnson2018multicalibration} learn a predictor for numerous overlapping demographic subgroups with a focus on calibration, and \citet{foulds2018intersectionality} propose an intersectional fairness regularizer that targets statistical parity. \citet{morina2019intersectional} propose a post-processing approach that generalizes that of \citet{hardt2016equalopp}, and \citet{kim2019multiaccuracy} similarly propose a post-processing approach as well as auditing procedure. \citet{friedler2019comparative} compare existing fairness methods, and consider intersectional sensitive attributes by encoding one axis of identity as race-sex.\footnote{One way of incorporating intersectional identities is by encoding them as, e.g., race-sex, such that \{\group{Black}, \group{White}\}$\times$ \{\group{Female}, \group{Male}\} can be considered as a single axis of identity with four values of \{\group{Black} \group{Female}, \group{Black} \group{Male}, \group{White} \group{Female}, \group{White} \group{Male}\}.}

Perhaps the most well-known of these works~\cite{kearns2018gerry,hebertjohnson2018multicalibration} never use the word ``intersectionality'', instead opting for the terms ``fairness gerrymandering'' and ``computationally-identifiable masses.'' Both works make important and impressive technical progress in generalizing algorithms for the intersectional setting, but by not explicitly naming ``intersectionality'', do not invoke the history, context, and literature that it brings.
\olga{yikers! they're going to be our reviewers, and primary target audience. tone down a lot.}

\section{Setup}
\label{sec:setup} 
Throughout our work, we provide experiments and empirical results to substantiate the claims we make.
In this section, we give an overview of the datasets, training objective, and algorithms that we perform such studies on. When faced with a choice to make about our experimental setup, e.g., which fairness metric to optimize for, we simplistically opt for the most straightforward choice that is most aligned with prior work in the space. This is because the goal of our work is not for exhaustivity in showing these issues will arise in \textit{every} fairness setting, but rather, 
that they do manifest in a generically adapted fairness setting with common algorithms trained on actual datasets.\footnote{Code is located at \url{https://github.com/princetonvisualai/intersectionality}}

\smallsec{Datasets}
We use the newly proposed tabular datasets derived from US Census data by \citet{ding2021retiringadult}. We do this because of both the reasons delineated by \citet{ding2021retiringadult}, such as the community's over-reliance on the Adult Income dataset~\cite{Dua:2019}, and also the richer data features available to us. 
For each dataset, we are able to query for additional demographic features for each individual, such as marital status and granular race labels, as needed.

We use the five datasets offered by the paper: ACSIncome, ACSPublicCoverage, ACSMobility, ACSEmployment, and ACSTravelTime.
We pick the California 2018 slice of these datasets to strike a balance between a computationally feasible size, and also having sufficient data points. This choice is somewhat arbitrary because, as we noted, we are not trying to make any data-driven societal insights, but merely demonstrate that particular phenomena may manifest in algorithms trained on actual datasets.
We assume the positive label of each dataset is the desirable one, even though this is not always clear, e.g., the positive label in ACSTravelTime corresponds to an individual traveling more than 20 minutes to get to work. However, we could conceive of a perhaps contrived setting in which getting predicted to have a longer travel time entails receiving some kind of travel stipend. Again, for the same reason as our selection of data slice, we do not place much weight into what would, in an application-based design, typically be very value-laden choices.

For all of our experiments, we perform five trials of each run, using random seeds and different training/validation/test splits for each, as recommended by \citet{friedler2019comparative}, to give 95\% confidence intervals.

\smallsec{Training Objective}
\label{sec:train_obj}
We train all algorithms to achieve a balance between measures of accuracy and group fairness.

Our measure of accuracy is \emph{soft accuracy}. Prior works have shown fairness metrics to be extremely sensitive to the classification threshold used~\cite{chen2020threshold}; hence we do not binarize the outputs, acknowledging that binarization may need to be done at application time to make direct predictions. For all $n$ individuals, let $y_i\in\{0, 1\}$ be the label for individual $i$, and $p_i\in[0, 1]$ be the probabilistic prediction for individual $i$. Soft accuracy is defined to be $\frac{1}{n}\sum_{i=1}^n y_i\cdot p_i+(1-y_i)\cdot(1-p_i)$.

Picking a fairness metric is highly non-trivial, as context about the downstream effects of the algorithm is needed.
However, for the scope of our work since we consider the positive labels to be more desirable, we choose a metric analogous to equal opportunity~\cite{hardt2016equalopp}, i.e., equalizing the true positive rate (TPR).\footnote{We focus on equal opportunity, but our flavor of analysis applies to other algorithmic fairness notions, such as demographic parity, equalized odds, etc.} 
Our measure of fairness is thus \emph{max TPR difference}.
To generalize the equal opportunity measure to more than two groups, we adopt a method similar to prior work~\cite{foulds2018intersectionality, kearns2018gerry, yang2020overlap}. If we define $TPR(g)$ to be the average $p_i$ for all individuals of group $g$ with label $y=1$, then our measure is the maximum pairwise difference between any two groups.
Most proposed fairness algorithms are able to optimize for this metric, and we are trying to capture the canonical way the community has been targeting intersectionality.
We will go on to investigate the sufficiency of metrics like this in Sec.~\ref{sec:evaluation}, and propose constructive suggestions there.\looseness=-1

For hyperparameter tuning we optimize for the geometric mean of \emph{soft accuracy} and (1 - \emph{max TPR difference}) to account for values with different scales. 

\smallsec{Algorithms}
\label{sec:algorithms}
Our experiments are performed on one baseline and five fairness algorithms.
Our baseline is a 3 layer fully connected neural network with 30 neurons in each hidden layer and a sigmoid activation trained to predict $y_i$ from an individual's features and demographic attributes. The first two fairness algorithms are general ones we extend to the intersectional setting by coding attributes as, e.g., race-sex: RWT~\cite{jiang2020reweight} is a reweighting schema and RDC~\cite{agarwal2018reductions} reduces to a sequence of cost-sensitive classifications. The latter three are intersectional methods: LOS~\cite{foulds2018intersectionality} has an extra intersectional fairness loss term, GRP~\cite{yang2020overlap} is a probabilistic combination of models, and GRY~\cite{kearns2018gerry, kearns2018empirical} produces cost-sensitive classifications from a 2-player zero-sum game. Details and hyperparameter search spaces are in Appendix~\ref{app_alg_and_hyper}.\looseness=-1

\section{Selecting which identities to include}
\label{sec:heterogeneity}

The first of three core challenges in incorporating intersectionality that we address in this work is considering which identities to include~\cite{mitchell2021prediction}.\footnote{We take as a given that identities should be included during training, i.e., fairness through awareness~\cite{dwork2012awareness, dwork2018decoupled}.} 
The foundation of this problem is that categorizing people into discrete, socially constructed groups, while tenuous, is often necessary for machine learning systems to make sense of socially relevant distinctions~\cite{edlagan2016disaggregate, hanna20race, jacobs2021measurement}. 
However, this flattening of individuals is often at the expense of ignoring different amounts of heterogeneity within each group. Homogeneity here would entail that each member of a group is best treated identically to all other members of that group by a machine learning model; heterogeneity involves a break from this assumption.\footnote{Because we have scoped our work to be on algorithmic harms, our investigation will be focused on heterogeneity's role in the context of model predictions. Thus, we will not perform what might be considered a more model-agnostic approach of unsupervised learning on the dataset itself.} 
In other words, one conception of a heterogeneous group is when, within that group, ``statistical patterns that apply to the majority might be invalid within a minority [sub]group''~\cite{hardt2014big}.
While heterogeneity will exist in any categorization of people, our focus is on the differing amounts of within-group heterogeneity that exists across groups.
A variety of machine learning approaches overlook this fact by assuming a version of constant within-group heterogeneity, whether that be through known variances across groups for a variational Bayesian approach to one-shot learning~\cite{feifei2003bayesian}, or the homoscedasticity assumption (i.e., that all groups have the same variance) for methods like linear regression and linear discriminant analysis. Violations of the homoscedasticity assumption are well-studied by statistical tests~\cite{glass1972consequences, lix1996assumption}, but less understood in the context of training machine learning models.\footnote{Differing heterogeneity is related to ``second moment'' statistical discrimination in economics: marginalized groups, for structural reasons like not being given sufficient opportunity to demonstrate ability, have a higher perceived variance and are discriminated against by risk-averse employers~\cite{aigner1977discrimination, england1992worth, dickinson2009statistical}.\looseness=-1}

The solution is not as trivial as simply adding in as many axes and granular identities as we have access to, which is also what often leads to outcries of how intersectionality might take us to the extreme of sub-dividing until each group is an individual person.
To demonstrate how we should consider which identities to include, we perform representative case studies on two racial groups. In Sec.~\ref{sec:hetero_api} we investigate the granularity of constituent identities within \group{Asian} \group{Pacific} \group{Islander} to include as labels (e.g., \group{Hmong}, \group{Japanese}, \group{Cambodian}, \group{Asian} \group{Indian}) in order to empirically explore the tension between adding more identities and reaching a point of intractability because there are too many groups. 
In Sec.~\ref{sec:hetero_other} we look into another heterogeneous racial group, \group{Other}, because how we go about including this category remains an important and relevant consideration so long as we are utilizing discrete categories.

One might ask why our case studies look into multiple groups within the same axis, rather than along different ones.
Different levels of heterogeneity within groups often come about due to additional axes of identity that are unaccounted for, e.g., gender differences within a racial group. However, we argue that heterogeneity along the same axis is also relevant, and an investigation of this will help us understand how to handle the intersectional case. 
We note that while socially the concepts of heterogeneity either due to additional axes or along the same axis are very different, technically they may warrant similar approaches. 
When along the same axis, the groups with higher heterogeneity are sometimes those that have been unified not because they share a particular trait, but rather because they share a hardship that has motivated them to pursue change as a more unified group, e.g., coalitional identities like \group{Disability}~\cite{adams2015disability, whittaker2019disability}. 
Another group likely to have high heterogeneity is \group{Other}, the residual group that comes with discrete categories. For example, if gender categories are \group{Male}, \group{Female}, and \group{Other}, this latter group may encompass people who identify as non-binary, intersex, and other gender identities that may differ greatly from each other.

\subsection{Case study: heterogeneity within \group{Asian} \group{Pacific} \group{Islander}}
\label{sec:hetero_api}
To investigate the granularity of identities to include, we consider ``Asian Pacific Islander'', or API. This racial grouping came about in the late 1960s, inspired by the Black Civil Rights Movement, as part of an initiative to unify disparate groups~\cite{api}. This aggregate category was on the US Census in 1990 and 2000, though a 1997 mandate separated it into ``\group{Asian}'' and ``\group{Native} \group{Hawaiian} \group{and} \group{Other} \group{Pacific} \group{Islander}.'' However, these two groups are frequently still clustered together, despite the very different forms of discrimination and stereotyping that each faces.

To understand the difference that label granularity of \group{API} makes, we perform a series of experiments where each algorithm is provided the same set of data, but with demographic features, $g$, along three different sets of granularity: ``\group{Asian}'' and ``\group{Native} \group{Hawaiian} \group{and} \group{Other} \group{Pacific} \group{Islander}'' are considered one aggregate group (1 group), ``\group{Asian}'' and ``\group{Native} \group{Hawaiian} \group{and} \group{Other} \group{Pacific} \group{Islander}'' are separated (2 groups), each group is further broken down into even more granular labels, such as \group{Hmong}, \group{Cambodian}, \group{Asian} \group{Indian}, etc (> 10 groups).
We preprocess the datasets to only include individuals with racial groups \group{White}, \group{Black}, and the granular groups within \group{API} with at least 300 individuals, 30 negative labels, and 30 positive labels. This is so as to not add too many variables for this particular case study.\looseness=-1

On the two datasets of ACSIncome and ACSTravelTime, we train and perform inference for each of our algorithms under the three granularity scenarios.
We use as our evaluation metric the max TPR difference, and always measure this between the most granular constituent groups we have, i.e., the scenario with > 10 groups. In other words, an algorithm considering \group{API} to be one aggregate group would be trained with these labels, and perform predictions with them. However, when evaluating in this setting, the more granular labels of > 10 groups are used. 

\begin{figure*}[t!]
  \centering
  \begin{minipage}[b]{0.98\textwidth}
    \includegraphics[width=\textwidth]{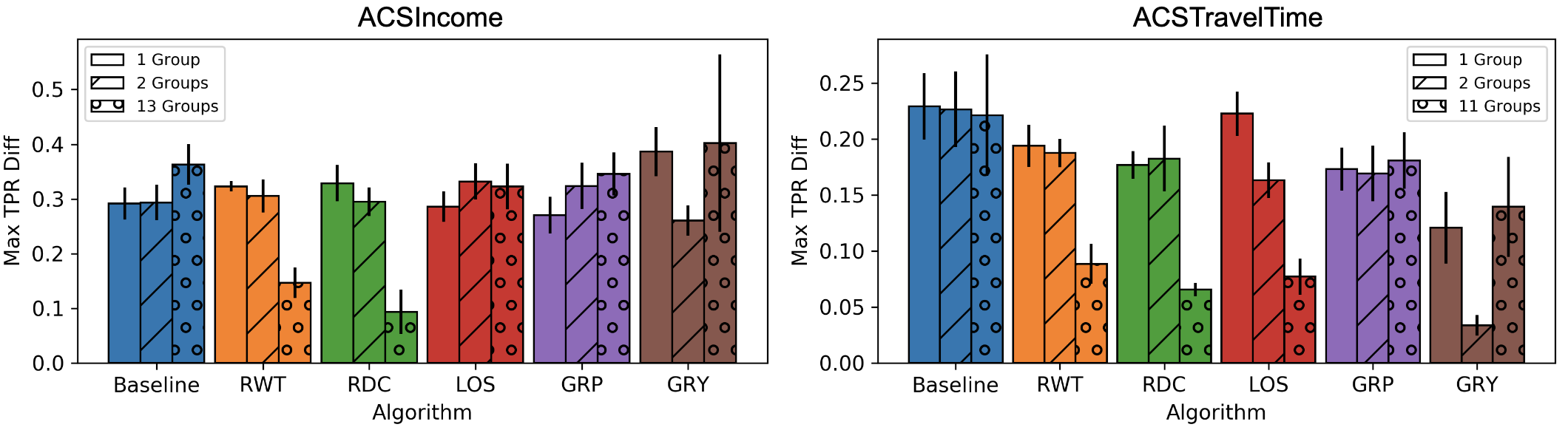}
  \end{minipage}
  \caption{Each algorithm is trained under three scenarios, where the \group{API} group is broken up into 1, 2, or > 10 granular groups. All algorithms are evaluated on the most granular setting of > 10 groups, where max TPR difference is only between these particular subgroups (excluding the \group{Black} and \group{White} racial groups). While on some dataset and algorithm combinations it is best to give the algorithm the most granular set of racial labels, in others it can actually benefit to use a more coarse set of labels.}
  \label{fig:hetero_api}
  \Description{Two graphs with the same setup on two different dataests of ACSIncome and ACSTravelTime. X-axis has six algorithms: Baseline, RWT, RDC, LOS, GRP, and GRY. Y-axis is the max TPR difference. Results are shown for each algorithm on 1 group, 2 groups, and more than 10 groups. Both graphs show conflicting results in terms of which granularity of group results in the most fair algorithm.}
\end{figure*}

Intuitively, one might posit that it is always best to use the most granular labels of > 10 groups when training, since these are the labels for which evaluation will be performed. However, that intuition breaks down when we consider empirical results in Fig.~\ref{fig:hetero_api}. Across both datasets for each algorithm, it is not always the case that training with the most granular labels results in the lowest max TPR difference for these very same groups. In fact, in the ACSTravelTime dataset, we actually see that GRY outperforms all other algorithms with a max TPR difference of $0.03\pm0.01$ when it considers \group{API} at the granularity of 2 groups. Possible reasons include that models may overfit when groups are small,  or that for some datasets, certain groups are sufficiently homogeneous that they benefit from being treated as the same. It is not always clear a priori which grouping is best for training, and this will require a combination of contextual understanding of the historical and societal reason behind the groupings in a particular domain as well as empirical validation to understand what works best in a particular setting. While in this case study our experimental scenarios were different granularities along a singular axis of identity, similar experiments can be performed where each scenario is the inclusion of a different combination of axes of identity.

\subsection{Case study: heterogeneity within \group{Other}}
\label{sec:hetero_other}
Considering how to handle the individuals that fall outside the delineated identities is an important consequence that comes with expanding beyond binary attributes or single axes of identity.
For example, multi-racial and non-binary individuals are often forced to pick a category that does not apply to them, or simply choose the catch-all \texttt{Other}, both of which have associated harms~\cite{hamidi2018agr, mclemmore2015misgender, ansara2012cisgenderism, scheuerman2021nonbinary, bowker2000classification, scheuerman2019gender}.
The \group{Other} racial category appeared in the US Census in 1910, and in 2010 was the third-largest racial category~\cite{ashok2016other}. Given that this group is defined by not belonging to any of the named racial groups, we might wonder if there is a larger amount of heterogeneity amongst the people who check this box. 
However, for race specifically, prior work has found that \group{Other} is not simply a leftover residual group, but rather a ``socially real phenomenon''~\cite{brown2007other}. 
For example, \citet{brown2007other} found that in the 2000 United States census, the \group{Other} racial group became a proxy for Hispanic people, where 97\% of people who checked ``Other'' for race also checked ``Hispanic'' for origin. 
Drawing from this, one might imagine that when training a model on a dataset exhibiting this characteristic, it could make sense to treat \group{Other} as its own group, or split up \group{Other} into those that are Hispanic or not.\looseness=-1

In the quantitative social sciences, there are three common ways in which racial groups like \group{Other} and \group{Multiracial} are approached: treating it as its own group (Separate), redistributing each individual to another group they are similar to~\cite{liebler2008multiplerace} (Redistribute), or simply ignoring this group~\cite{froharddourlent2016options} (Ignore). We empirically test these three approaches to understand the different accuracies they result in for the \group{Other} group. To implement Redistribute, we simplistically pick a strategy of re-assigning each individual in \group{Other} to the racial group of its nearest L2 distance neighbor in the feature space. While results from this method of redistribution may not generalize to other methods, we show this as one demonstrative example.
The reason we might believe this strategy could be helpful is if individuals in the \group{Other} group have distributions such that being grouped with another group would lead to more accurate predictions. 



\begin{figure*}[t!]
  \centering
  \begin{minipage}[b]{0.9\textwidth}
    \includegraphics[width=\textwidth]{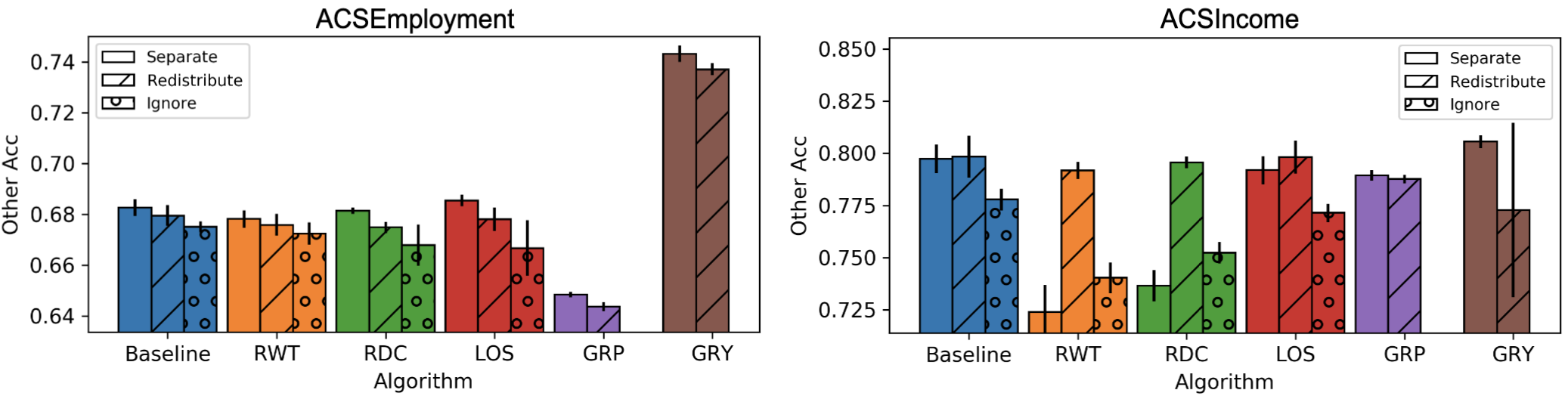}
  \end{minipage}
  \caption{Three methods of incorporating the \group{Other} racial group: Separate - treating \group{Other} as its own group, Redistribute - redistributing each member of \group{Other} to the racial group of its nearest neighbor, and Ignore - ignoring this group at train time. GRP and GRY are unable to perform inference on groups unseen at training time, so those results could not be generated. For ACSEmployment, Separate is often the best condition. However, for ACSIncome, no condition is best for all scenarios.}
  \label{fig:hetero_other}
  \Description{Two graphs with the same setup on two different datasets of ACSEmployment and ACSIncome. X-axis has six algorithms: Baseline, RWT, RDC, LOS, GRP, and GRY. Y-axis has the AUC for the Other racial group. Results are shown for each algorithm on Separate, Redistribute, and Ignore. Left graph for ACSEmployment shows Separate to be the best condition for most algorithms. Right graph for ACSIncome shows different conditions to perform well, such as Redistribute over Separate for RWT and RDC.}
\end{figure*}

Similar to the case study on \group{API}, we train our model across three different scenarios while keeping the evaluation consistent. 
In Fig.~\ref{fig:hetero_other} we see that for ACSEmployment, the Separate scenario of treating \group{Other} as its own distinct group does frequently perform best---not necessarily unexpected given our prior domain knowledge. However, in Fig.~\ref{fig:hetero_other} for ACSIncome it is no longer always the case that treating \group{Other} as its own group performs better, e.g., Redistribute outperforms Separate for the RWT and RDC algorithms. This furthers our finding that contextual knowledge, e.g., about \group{Other} sometimes being a racial group in its own right, is not entirely sufficient in helping us know a priori how best to handle a group.\looseness=-1

\subsection{Constructive Suggestions}
\label{sec:hetero_suggest}

Overall, our case studies show it is rarely clear a priori which identities should be included due to differences in performance on even the same task across algorithms.
Ultimately, it will take a \suggest{combination of contextual understanding of the application and empirical experiments} to make the decision on what works best for a particular application.
When \citet{crenshaw1989intersectionality} first introduced intersectionality, she was considering it in the context of how race discrimination law and sex discrimination law failed to capture the discrimination experienced by Black women. Thus, the relevant axes for her to consider were race (\group{Black} and \group{White}) and sex (\group{Female} and \group{Male}). Guided by this kind of contextual knowledge, practitioners can incorporate empirical findings to select the relevant set of identities to include.

\suggest{When training predictive models, including all available identity labels may not always be best.} Computational tractability comes into play, as do similarities and differences between groups. For example, if group \group{A} and \group{B} have similar distributions, but group \group{B} has a small number of labels, it may benefit this group to join with group \group{A}, lest there not be sufficient samples to train a predictive model. The exception is for post-processing approaches, where it may be more likely to be beneficial to include more identities. 
This is because these approaches generally involve learning only one or two group-specific parameters, e.g., classification threshold or probability of flipping a binary prediction~\cite{hardt2016equalopp, pleiss2017calibration, kamiran2012decision}, and are thus less susceptible to overfitting.

Another suggestion regarding which identities to include is on how to handle the individuals that fall outside the delineated identities. 
Experiments guided by contextual knowledge show us how to proceed. 
However, we advise extreme caution when considering any kind of recategorization due to serious normative implications. 
\citet{benthall2019categories} proposed using unsupervised clustering to discover ``race-like'' categories, and \citet{hanna20race} levied a critique against such a mechanism. 
Use of unsupervised recategorization without human oversight can lead to inherently harmful actions, such as recategorizing someone who identifies as \group{Non-Binary} to a gendered grouping they do not belong to. 
However, supplementing recategorization with domain expertise, such as with the \group{Other} racial group, may be considered more permissible.
As another option, \citet{hancock2007multiplication} has proposed fuzzy set theory to better capture the categorization of people into socially constructed categories, and \citet{mary2019continuous} provides technical guidance on this.

In contrast, \suggest{for evaluation, one should generally perform analyses at the most fine-grained level possible.} In fact, in many cases if a dataset does not contain sufficient axes or identities according to a domain expert's understanding of the kinds of intersectional power dynamics that may be at play, one should consider collecting more demographic labels, keeping privacy concerns in mind~\cite{jo2020archives}. 
For smaller groups where one might argue evaluations would be noisy~\cite{foulds2019measure}, error bars could convey uncertainty, and evaluations on the larger group could also be included.

\section{Handling progressively smaller groups}
\label{sec:dataset}
Now that we have discussed which identities to include, we consider the next inevitable challenge that comes with incorporating more identities: the presence of ever smaller numbers of individuals in each group. Machine learning has long tackled such long tail problems, but there are a few critical distinctions. One notable technical one is the long tail that machine learning concerns itself with is typically \textit{in} the label, whereas in intersectionality it is \textit{within} the label. In other words, it is not that there are too few examples of chairs, but rather a small set of the many chairs are wooden and look different from the others. 
Although this difference is important to keep in mind, it is rather the normative distinctions we will discuss that may impact the transferrability of existing data imbalance techniques to tasks with intersectionality concerns.
However, this also gives rise to a possible new, underexplored approach that leverages the structure of intersectional data. 
We first walk through a set of traditional dataset imbalance techniques in machine learning in Sec.~\ref{sec:dataset_transferrence}, and then present the kind of structure that may provide such a new avenue in Sec.~\ref{sec:dataset_structure}.


\subsection{Dataset imbalance techniques in machine learning}
\label{sec:dataset_transferrence}

A common machine learning technique for dealing with imbalanced classes is simply reweighting or resampling~\cite{kubat1997imbalanced, japkowicz2000imbalance}. 
These entail simply increasing the attention paid to individuals of a particular group. 
Reweighting can also be done in an adaptive and learned way~\cite{ren2018reweight, jiang2020reweight}. This might pose a normative problem because now we leave up to the model's learning parameters which individuals will be over- or under-valued.

Further techniques to tackle class imbalance move from changing the importance of existing training samples to generating new synthetic examples. 
Techniques like SMOTE~\cite{chawla2011smote} generate synthetic examples in the feature space, and there is no immediately clear intuition on the normative concern of injecting this kind of directed noise to an abstract space of features.
There is also a vast literature discussing counterfactuals, a type of synthetic example steeped in causality, and which have long been used to detect forms of discrimination~\cite{chiappa2019causal, li2017detection}. Their permissibility of use, however, for addressing class imbalance has been contested because of the infeasibility of manipulating demographic categories~\cite{holland2008causation, sen2016sticks, kasirzadeh2021counterfactuals} and inaccurate conceptualizations of causality of demographic categories~\cite{kohlerhausmann2019counterfactual, hu2020sex}.

When we move from the data space of abstract features to one like images of faces, manipulations often feel viscerally wrong, e.g., Figure 1 of \cite{yucer2020face}. Recent work has proposed leveraging Generative Adversarial Networks (GANs) to create synthetic examples to help train models, especially for facial datasets~\cite{ramaswamy2021latent, sattigeri2019gan, sharmanska2020contrastive},\footnote{We leave out of scope the concerns with facial recognition itself, pointing to works like~\cite{stark2019plutonium, goldenfein2019profiling, petty2020surveil}} but some of these visual results can feel akin to the harmful performance of blackface. This poses a set of questions, such as if such generations were to actually help train a model, and at the cost of less privacy concerns that might result from seeking to explicitly, and perhaps exploitatively, collect more data of underrepresented groups, is there anything to be gained from it? If we consider the normative concern to be the harmful visibility of these images because of their historical context, might we consider the generation of these images permissible, so long as they are only to be consumed by a machine learning model? These remain open questions.

One field in which synthetic examples are being widely adopted, and where results are relatively accepted as indicative of real-world gains largely due to work on transfer learning (e.g., Sim2Real~\cite{sim2realworkshop}), is reinforcement learning for robotics~\cite{kober2013rlsurvey}.
Simulation can often provide a safer alternative to training in the real world, and serve as a source for more data.
However, because the data needs of certain fairness applications, like in tabular domains, are lower than that of reinforcement learning applications, which may use image data, the use of simulation for fairness applications may better serve a different purpose.
For example, modeled after OpenAI's Gym~\cite{openaigym} (a simulation testbed for reinforcement learning), FairGym has been proposed to explore long-term impacts of fairness problems~\cite{damour2020fairgym}. Thus, like \citet{schelling1971segregation} did to study segregation, the purpose of simulations for problems with intersectional concerns may be more akin to that of simulations as testbeds and tools for understanding sociotechical systems such as in modeling recommender systems~\cite{chaney2018recommendation, lucherini2021trecs, ciampaglia2018popularity, geschke2018filter, jiang2019degenerate} and online information diffusion~\cite{garimella2017balancing, goel2016virality, tornberg2018echo}, rather than as a source for more data.

\subsection{Structure in data}
\label{sec:dataset_structure}
One avenue that exists in intersectionality for handling progressively smaller subgroups is leveraging the structure of the dataset. 
Drawing on our previous notion of homo- and heterogeneity, we consider that when two different groups are similar, there may be predictive patterns we can learn about one from the other. We deviate slightly from our previous notion by ignoring changes in base rate to focus only on changes in the mapping of the input distribution to output distribution.\footnote{We ignore base rate differences because these are easier to account for using post-processing approaches that learn a minimal number of parameters.}
Thus, in this section, we use ROC AUC as our evaluation metric because it is base rate agnostic, allowing us to focus on what we call ``predictivity'' differences. One major concern that the methods in the previous Sec.~\ref{sec:dataset_transferrence} aimed to address is that underrepresentation can be a problem if the minority group is differently predictive from the majority group~\cite{larrazabal2020imbalance}, but does not contain sufficient samples to train a robust model on. Structure in the data has the potential to help us alleviate this concern if we can learn something about the predictivity of an underrepresented group, e.g., \group{Black} \group{Female}, from groups with more representation in the dataset and a shared identity, e.g., \group{Black} \group{Male} who share the attribute of \group{Black}. It remains important of course to consider how the context might impact the structure available to draw from. For example, \group{Gay} \group{Female} and \group{Gay} \group{Male} may not share predictivity, despite sharing the attribute of \group{Gay}~\cite{monteflores1978comingout,herek2002gay}.\looseness=-1

Two kinds of predictivity difference are of relevance to us: between subgroups (i.e., each subgroup is differently predictive from each other) and an additional intersectional effect (i.e., the predictivity of one group cannot be learned from groups with which it shares identities). The presence of subgroup predictivity differences tells us we should be concerned with underrepresentation, and a lack of intersectional predictivity differences tells us we may be able to leverage structure in the data to alleviate this. We will perform two experiments, each aiming to discover one type of predictivity difference.\looseness=-1

We study three datasets: ACSIncome, ACSMobility, and ACSTravelTime to understand the types of predictivity difference present in each. We consider the demographic attributes \{\group{Black}, \group{White}\}$\times$ \{\group{Female}, \group{Male}\}. We focus on the group \group{Black} \group{Female} to center their experience, and because this group is the most underrepresented across these datasets and thus more likely to face underrepresentation. We perform these experiments using the Baseline model.

We first investigate subgroup predictivity differences. To do so, we train only on individuals from one of the four groups at a time, controlling for the number of training samples to be constant, and test on \group{Black} \group{Female}. 
For example, by comparing the AUC of \group{Black} \group{Female} when trained on $n$ samples of \group{White} \group{Female} as compared to when trained on $n$ samples of \group{Black} \group{Female}, we can understand whether there is a predictivity difference between the two. Our results across the three datasets are in Tbl.~\ref{tbl:dataset_predictive}, where we can see that while for ACSIncome the model is able to achieve roughly the same AUC on \group{Black} \group{Female} no matter the group it was trained on, for ACSPublicCoverage and ACSTravelTime, \group{Black} \group{Female} has the highest AUC when a model is trained on members from the same group, rather than a different group. This indicates that in these two datasets, \group{Black} \group{Female} is differently predictive from other groups such that any model without sufficient samples of \group{Black} \group{Female} may not perform as well, and thus an underrepresentation of \group{Black} \group{Female} training samples may be of concern. This is not true in ACSIncome, since an underrepresented group like \group{Black} \group{Female} shares predictivity with other groups for which there are sufficient training samples.

\begin{table*}
\caption{For the three datasets, \group{Black} \group{Female} AUC with 95\% confidence interval is shown when trained on only one subgroup at a time. By comparing AUC across different training subgroups, differing predictivities can be seen. For ACSIncome, the predictivity of \group{Black} \group{Female} does not differ much from other groups, but this is not true for ACSPublicCoverage and ACSTravelTime.}
\label{tbl:dataset_predictive}
\centering
\begin{tabular}{>{\centering\arraybackslash}p{2.6cm}|>{\centering\arraybackslash}p{2.4cm} |>{\centering\arraybackslash}p{2.4cm} >{\centering\arraybackslash}p{2.4cm} >{\centering\arraybackslash}p{2.4cm}}
\toprule
 Dataset & Black Female & White Female & Black Male & White Male \\ 
 \hline
 ACSIncome & $75\pm2 $ & $77\pm2$& $75\pm1$&$76\pm1$ \\  
 ACSPublicCoverage &\textbf{74}$\pm$\textbf{1} & $71\pm1$ & $73\pm2$ & $72\pm2$ \\
 ACSTravelTime & \textbf{67}$\pm$\textbf{2} &  $63\pm2$& $64\pm2$ & $59\pm1$ \\
 \bottomrule
\end{tabular}
\end{table*}

\begin{figure*}[t!]
  \centering
  \begin{minipage}[b]{0.98\textwidth}
    \includegraphics[width=\textwidth]{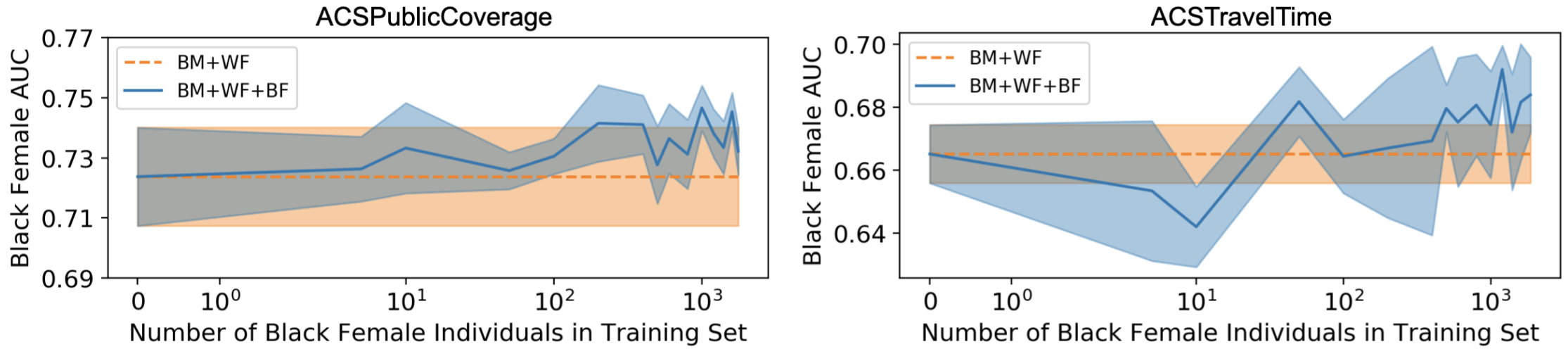}
  \end{minipage}
  \caption{For two datasets that exhibit subgroup predictivity difference, we investigate the intersectional predictivity difference. The orange dashed ``BM+WF'' line represents the highest \group{Black} \group{Female} (BF) AUC achievable when only trained on a combination of \group{Black} \group{Male} (BM) and \group{White} \group{Female} (WF). The blue solid line ``BM+WF+BF'' represents the BF AUC when samples of this particular group are added into the training set. Shaded region indicates 95\% confidence interval. Training samples are controlled for such that every data point has been trained on the same number of samples. The left graph shows that replacing individuals from BM+WF with BF does not increase the AUC of BF very much, indicating there is likely no intersectional predictivity difference. The right graph shows the addition of BF individuals does increase the AUC of BF, indicating the presence of an intersectional predictivity difference.}
  \label{fig:dataset_bf}
  \Description{Two graphs for ACSPublicCoverage and ACSTravelTime datasets. X-axis is "number of Black Female individuals in training set." Y-axis is "Black Female AUC." Left graph shows two lines with overlapping confidence intervals. Right graph shows two lines with overlapping confidence intervals until the end, when the "BM+WF+BF" line is slightly higher than "BM+WF."}
\end{figure*}

We are now faced with the finding that in ACSPublicCoverage and ACSTravelTime, the group \group{Black} \group{Female} is differently predictive such that a model trained without sufficient examples of this group will not perform as well.
To address this concern, we can consider the known structure in intersectional data that can be leveraged. In other words, there may be something about the predictivity of \group{Black} \group{Female} that we can learn from \group{Black} \group{Male} or \group{White} \group{Female}. To understand the limit of this structure, and the extent to which there is an \textit{extra} intersectional effect whereby there remains predictivity differences about \group{Black} \group{Female} that cannot be learned from the groups of \group{Black} and \group{Female}, we consider a different experiment. We operationalize the limit of what can be learned from the groups that share identities with \group{Black} \group{Female}, without training on any members from this group itself, by picking the ratio of \group{White} \group{Female} to \group{Black} \group{Male} training samples that results in the highest AUC performance on a validation set of \group{Black} \group{Female}. We perform this experiment for ACSPublicCoverage and ACSTravelTime, and in Fig.~\ref{fig:dataset_bf} (left) see that for ACSPublicCoverage, while \group{Black} \group{Female} has a unique subgroup predictivity, this difference can largely be learned from groups with shared characteristics, i.e., \group{White} \group{Female} and \group{Black} \group{Male}. On the other hand, in Fig.~\ref{fig:dataset_bf} (right) for ACSTravelTime, training only on groups that share characteristics with \group{Black} \group{Female} results in a AUC lower than what can be achieved when trained on the actual group tested upon. 
Although these results are somewhat noisy, they suggest that for this dataset and model, there may be an extra intersectional predictivity unique to \group{Black} \group{Female}.

Thus, our two experiments tell us the following: ACSIncome - no subgroup predictivity difference, ACSPublicCoverage - subgroup but not intersectional predictivity difference, and ACSTravelTime - both a subgroup and intersectional predictivity difference. Based on these results, ACSPublicCoverage appears eligible for leveraging dataset structure to alleviate an underrepresentation of the \group{Black} \group{Female} group.
This means that algorithms need to incorporate additional axes of identities in a more structured way than just encoding the conjunction of attributes, e.g., race-sex. 
We emphasize that these results on predictivity do not make any claims about the societal causes of these effects.

\subsection{Constructive Suggestions}

In Sec.~\ref{sec:dataset_transferrence} we warn against transferring existing data imbalance methods to handle the progressively smaller groups that will arise in intersectionality without first giving careful consideration to normative concerns. In Sec.~\ref{sec:dataset_structure}, we provide an initial exploration into a direction that \suggest{leverages the structure in the data to learn about an underrepresented group from other groups it shares identities with}. We recommend performing data analysis like we have shown to demonstrate whether this may be a suitable approach to proceed with for a particular algorithm and task.

However, this approach comes with significant caveats. Relying on it too much can counteract the purpose of intersectionality, and the unique effects that multiply marginalized groups experience.
Prior work warns against treating intersectionality as a variable to be controlled for in an additive or multiplicative way~\cite{simien2007intersectionality, hancock2007multiplication}, and though our use of a multi-layer perceptron allows differences in predictivity to be captured in more complex ways than as a regression variable, we emphasize that mathematical structure in no way implies societal structure. For example, structure in the data does not preclude the idea that gender can only be understood in a racialized way, and vice versa~\cite{simien2007intersectionality}.

\section{Evaluating a large number of groups}
\label{sec:evaluation}
Finally, we consider the problem of evaluation, and how extrapolating existing metrics is insufficient when the number of subgroups considered increases. In the binary attribute setting, fairness evaluation frequently takes the form of the difference between the groups for some performance measure derived from the confusion matrix. When more than two groups are considered, the evaluation metrics look similar in that they are generalized versions, rarely with changes to account for the incorporation of additional groups. They are commonly formulated in terms of the maximum difference (or ratio) of a performance metric either between one group to that of all~\cite{guo2021emergent, kearns2018gerry, yang2020overlap}, or between two groups~\cite{foulds2018intersectionality, ghosh2021worst}. We go over problems with both conceptualizations, and then offer suggestions for additional metrics to measure---notably, that it is important to consider the relative rankings of demographic groups.

\subsection{Weaknesses of existing evaluation approaches}
\label{sec:evaluation_weaknesses}

One-vs-all metrics will far more frequently measure minority groups to have the highest deviation, because inherently the majority group has the most influence over what the ``all'' is. 
In the other conceptualization via pairwise comparisons, only the values of two groups, usually the maximum and minimum, are being explicitly incorporated while the rest are ignored. The more groups there are as a result of incorporating more axes and identities, the more values are ignored. For example, when using max TPR difference in a setting with three demographic groups, we might imagine two different scenarios such that in one, the three groups' TPRs are $\{.1, .2, .8\}$, and in another, they are $\{.1, .6, .8\}$. Both would report the same measurement of .7, despite coming from different distributions.

We term the family of evaluation metrics that encompasses both of these categories to be ``max difference.''
In line with existing work~\cite{foulds2018intersectionality, kearns2018gerry, yang2020overlap}, our max TPR difference defined in Sec.~\ref{sec:train_obj} is of this variety. However, there are significant problems with this formulation, even in the binary attribute setting. Because the mathematical notation for these fairness constraints is often formulated using parity (e.g., for groups $g_1$ and $g_2$, $TPR(g_1) = TPR(g_2)$), an absolute value is sometimes applied to the difference, which would obscure whether an algorithm has over- or under-corrected.

\begin{table*}
\caption{In ACSIncome, Group \group{A} has the lowest base rate, and Group \group{B} has the highest. After training to lower max TPR difference, all algorithms except GRP improve upon the baseline. However, all consistently rank Group B above Group A, reifying this hierarchy.}
\label{tbl:ranks}
\centering
\begin{tabular}{c| c |>{\centering\arraybackslash}p{2cm} >{\centering\arraybackslash}p{2cm}}
\toprule
 Algorithm & Max TPR Difference (\%) & Average Group A Rank & Average Group B Rank \\ 
 \hline


 Baseline & $13.9\pm2.9 $ & 4.0 & 1.0 \\  
 RWT & $2.8\pm0.4$ & 3.0 & 1.6 \\
 RDC & $3.0\pm1.2$ & 3.0 & 1.4 \\
 LOS & $4.7\pm1.8$ & 3.6 & 1.0 \\
 GRP & $18.2\pm1.2$ & 4.0 & 1.0 \\
 GRY & $6.7\pm1.1$ & 4.0 & 1.0 \\
 \bottomrule
\end{tabular}
\end{table*}

To demonstrate further shortcomings, we first consider the ACSIncome dataset and the demographic groups of \{\group{Black}, \group{White}\}$\times$ \{\group{Female}, \group{Male}\}. To get an idea of fairness concerns we miss by only considering a max difference metric like max TPR difference, we also calculate an additional metric that includes a notion of group rank. 
For Group A which has the lowest positive label base rate of the four groups, and Group B which has the highest, we report the ranking of their TPR (from 1 to 4 with 1 as the highest) relative to the other groups. In Tbl.~\ref{tbl:ranks} we see that across all algorithms, even if fairness is improved from the Baseline model, Group A's ranking is consistently low while Group B's is consistently high, and always higher than that of Group A's. This is crucial to know because despite a fairness criteria of max difference below some $\epsilon$ being satisfied, the consistent ranking of one group below another compounds in a way to further systematic discrimination~\cite{creel2021leviathan}. That there is a correlation between max TPR difference and group ranking in this scenario does not negate the importance of one metric or the other, as they each convey different information.

In our next experiment, we consider when there are many more than four subgroups and look at a new metric that measures the correlation between the rankings of a) base rates and b) TPRs after the fairness algorithm has been applied.
This ranking correlation helps us understand to what extent the underlying social hierarchy is upheld.\footnote{We note here a difference in what we are proposing from the syntactically similar space of fairness in rankings~\cite{singh2018exposure, zehlike2021rankingsurvey} There, those being ranked are individuals, and the goal is to more closely align to a set of ground-truth rankings. Here, those being ranked are the aggregate evaluation metrics of demographic groups, and the goal is to surface alignment to existing rankings as a way of providing additional understanding about a model.}
The higher the correlation between these two sets of rankings, the more we are reifying a particular hierarchy of subgroups and entrenching existing disparities in the data. We use Kendall's Tau~\cite{kendall1938tau} as a measure of rank correlation, and 
combine the p-values obtained across runs of random seeds using Fisher's combined probability test~\cite{fisher1925stats}. 

\begin{figure*}[t!]
  \centering
  \begin{minipage}[b]{0.95\textwidth}
    \includegraphics[width=\textwidth]{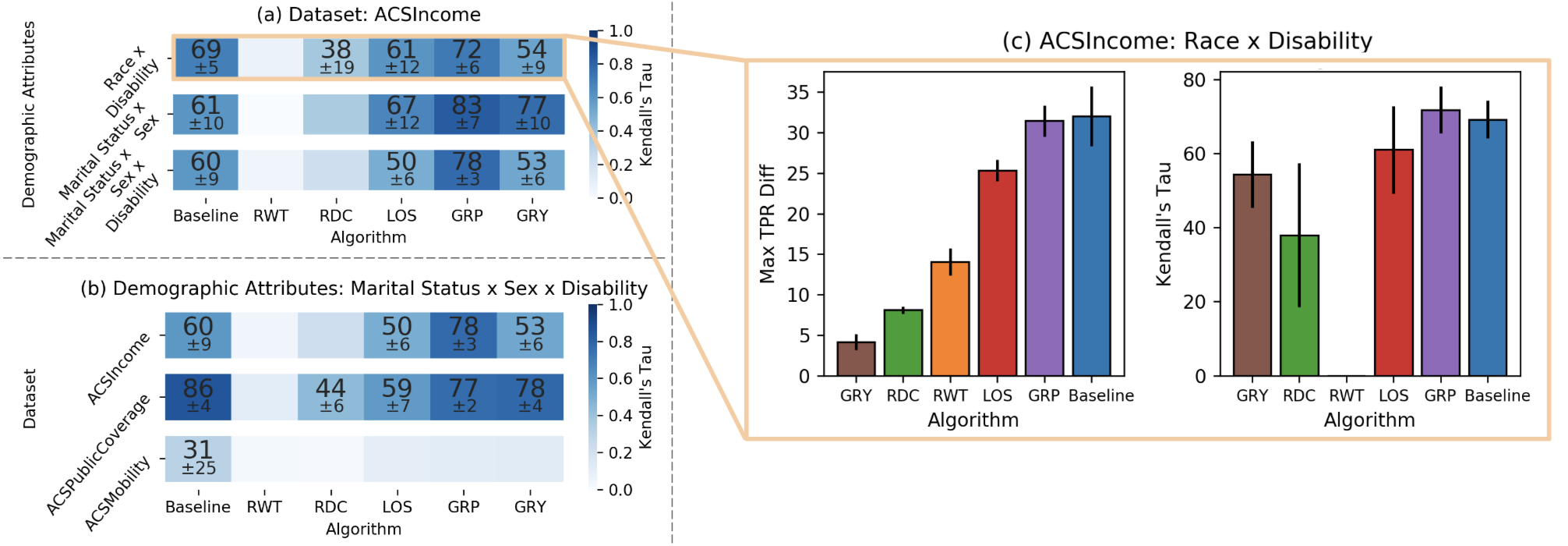}
  \end{minipage}
  \caption{In (a) we hold dataset constant and vary the axes of demographic attributes, and in (b) we hold demographic attributes constant and vary the dataset. We discern trends across algorithms (RWT and RDC are less likely to reify hierarchies compared to GRP and GRY), demographic attributes (for ACSIncome, marital status x sex has predictive patterns more likely to entrench hierarchies), and dataset (for marital status x sex x disability, ACSMobility has predictive patterns least likely to entrench hierarchies). 
  In (c) we display the first row of (a)'s max TPR difference and Kendall's Tau correlation values. We find that correlations between base rates and TPRss remain high across five of the six algorithms, and that the two metrics present different pictures of which algorithms perform best.
  }
  \label{fig:evalrank}
  \Description{Two graphs on the left report the Kendall's Tau value if it is statistically significant. The first holds dataset constant at ACSIncome and varies demographic attributes across: 1) race x disability, 2) marital status x sex, 3) marital status x sex x disability. The next holds demographic attributes constant at marital status x sex x disability and varies the dataset across ACSIncome, ACSPublicCoverage, and ACSMobility. On the right, a graph shows ACSIncome when demographic attributes are race x disability, and presents the max TPR difference and Kendall's Tau value across six algorithms. The rankings amongst the algorithms between the two metrics are different.}
\end{figure*}

Our results are in Fig.~\ref{fig:evalrank}, and we only display the Kendall's Tau value when it is statistically significant with $p<.05$. For the two graphs on the left, in Fig.~\ref{fig:evalrank}(a) we hold the dataset constant and vary the axes of demographic attributes, and in Fig.~\ref{fig:evalrank}(b) we hold the axes of demographic attributes constant and vary the dataset. 
We see that there are trends across algorithms (RWT and RDC are less likely to reify underlying rankings compared to GRP and GRY), demographic attributes (for ACSIncome, marital status x sex has predictive patterns more likely to reify underlying rankings), and dataset (for marital status x sex x disability, ACSMobility has predictive patterns less likely to reify underlying rankings). 

In Fig.~\ref{fig:evalrank}(c) we now take the setting from the first row of Fig.~\ref{fig:evalrank}(a) and compare the two metrics of max TPR difference and our Kendall's Tau ranking correlation. This shows two weaknesses with existing max difference metrics. The first, also demonstrated by Tbl.~\ref{tbl:ranks}, is that additional information about how closely a model's outputs adhere to underlying rankings provides an important and new perspective in understanding a model, as across five of the six algorithms, there is a statistically significant correlation in ranking.
The second, is that across-algorithm comparisons can lead to conflicting conclusions, as each metric conveys a different algorithm to be more ``fair.'' Under max TPR difference, GRY is best, whereas for ranking correlation, it is RWT.
These trade-offs need to be navigated by someone informed of the downstream application, and not implicitly ignored through the measuring of just one metric or another.\footnote{Across both experiments from Tbl.~\ref{tbl:ranks} and Fig.~\ref{fig:evalrank}, we are not comparing positive predictive value (PPV), but rather True Positive Rate (TPR), which is anchored in the $y$ labels as the ground truth. In other words, we are taking a rather conservative approach to fairness~\cite{foulds2018intersectionality}, because even in a scenario where TPRs are equal, PPVs could still exactly correlate with base rates. That even under this more conservatively fair conception the existing inequalities are reified so strongly, signifies one can only imagine how much larger the inequalities would be when considering PPV.}

The use of max difference metrics is emblematic of a larger trend in machine learning whereby all categories are generally treated the same. Although sometimes labels are treated differently, e.g., medical and self-driving car domains where FNs are more significant than FPs, this is a difference in the label rather than subgroup. 
It would make no difference to a model or evaluation metric if the labels for \group{Black} \group{Female} and \group{White} \group{Male} were swapped --- a surprising statement when considering intersectionality and the importance of the history of oppressed groups. As machine learning fairness begins to consider intersectionality, we need to resist evaluation metrics that do not substantively incorporate additional considerations, and merely extrapolate from existing metrics. This is not to say that max difference isn't useful, but rather that we should also consider others.\looseness=-1

\subsection{Constructive Suggestions}
We offer suggestions on being more thoughtful with pairwise comparisons, as well as additional types of evaluation.\footnote{Aggregating a set of input values into one summary output is akin to the economic framework of social choice theory. Social choice theory handles the aggregation of a set of individual’s inputs, which typically take the form of preferences, votes, welfare, etc.~\cite{arrow1951socialchoice}. Each individual has an input utility value under each possible world state, and an aggregation rule is chosen over the inputs in order to pick the best world state. We could imagine leniently conceiving of each individual in the set to be a different pair of demographic subgroups, with its corresponding utility being the negative TPR difference between them. The different world states are then the set of all possible model predictions. Under this conception, max TPR difference would map to the \textit{egalitarian} aggregation rule~\cite{rawls1974maximin}, which maximizes the minimum utility of all individuals (i.e., minimizing the max pairwise TPR difference). Under a different, \textit{utilitarian} aggregation rule~\cite{mills1863utilitarianism, moulin2004welfare}, we would instead maximize the sum of the utility values (i.e., minimize the sum of pairwise TPR differences). Ultimately we did not further explore this perspective because our focus is to expand out beyond this flavor of aggregation; however, one could imagine this to be a direction of exploration, e.g., by conceiving of individuals to be an entity other than a pair of demographic subgroups.}

If pairwise comparisons are to be done, machine learning practitioners can learn from other disciplines. When social scientists leverage pairwise comparisons in studying intersectionality, they often do so with more \suggest{deliberate thought put into the pairings, i.e., with a context-first rather than numbers-first approach} (e.g., between the max and min). As noted by \citet{mccall2005complexity}, ``although a single social group is the focus of intensive study, it is often shown to be different and therefore of interest through an extended comparison with the more standard groups that have been the subject of previous studies.'' 
The author gives examples like comparing ``working-class women to working-class men \cite{freeman2000highheels}'' and ``Latina domestic workers to an earlier generation of African American domestic workers \cite{hondagneusotelo1994domestica}.''
There has been work in machine learning that used a somewhat context-first approach in intersectional evaluations, but frequently default to anchoring comparisons to the most privileged group, e.g., white men~\cite{steed2021unsupervised, tan2019word}, or the least privileged group~\cite{tan2019word}. However, this can reify the norm of the privileged as default, a complaint that has been made about certain intersectionality frameworks~\cite{carastathis2008privilege}. \citet{johfre2021reference} note that even if done out of convention, comparing relative to a dominant group can reify the notion that they are the norm. The paper puts forth concrete guidelines on how to better choose the reference category, and though not specific to intersectionality, can help us navigate how to be more deliberate with any pairwise comparisons that need to be performed. While we point to their work for details, this includes heuristics such as if some group is defined as the negation of another or if certain categories unfold from one singular group.\looseness=-1

Reporting disaggregated analyses for all subgroups would of course alleviate many of these problems, and should be done before deployment~\cite{barocas2021disaggregate}, but for iterating on model training can be unwieldy. \suggest{Additional ``summary statistics'' that involve just adding one or two more bits of information}, such as the metrics we show in Sec.~\ref{sec:evaluation_weaknesses} of the ranks of the groups with the highest and lowest base rate or the correlation between the rankings of the base rates and model TPRs, would greatly supplant just the pairwise difference.
While each individual algorithm may seem fair, if each algorithm has group \group{A}'s TPR $\epsilon$ below group \group{B}'s, this can have significant compounding impacts on individuals from group \group{A}~\cite{creel2021leviathan}.\looseness=-1


\section{Conclusion}
In this work, we consider the problems that machine learning fairness will need to grapple with as it endeavours upon the process of incorporating intersectionality. We provide guidance on three practical concerns along the machine learning pipeline. For which identities to consider, we recommend evaluating on the most granular intersecting identities available in the dataset, but combining domain knowledge with experiments to understand which are best to include when training models. For how to handle the increasingly small groups, we caution against porting over existing machine learning techniques for imbalanced data due to their additional normative concerns, and offer a suggestion about leveraging structure that may be present between groups that share an identity. And finally for evaluation of a large number of subgroups, we both suggest how one could more thoughtfully conduct pairwise comparisons as well as present additional metrics to capture broader patterns of algorithms which existing metrics may obscure.
These are just a few of the many steps that will need to be taken to incorporate intersectionality into machine learning, and we encourage the machine learning community to grapple with the complexities of intersectionality beyond just conceptualizing it as multi-attribute fairness.\looseness=-1

\section{Positionality Statement}
All authors are computer scientists by training, and despite having worked on ML fairness, we do not have traditional social science backgrounds.
Additionally, there are group identities we discuss that we don't have lived experiences for.\looseness=-1


\begin{acks}
This material is based upon work supported by the National Science Foundation under Grant No. 1763642, Grant No. 2112562, and Graduate Research Fellowship to AW. Any opinions, findings, and conclusions or recommendations expressed in this material are those of the author(s) and do not necessarily reflect the views of the National Science Foundation.
We thank Susan J. Brison, Chris Felton, Arvind Narayanan, Vivien Nguyen, and Dora Zhao for feedback. 
\end{acks}



\bibliographystyle{ACM-Reference-Format}
\bibliography{bib}

\appendix
\section{Algorithms and Hyperparameters}
\label{app_alg_and_hyper}
We describe additional details about the five algorithms we perform experiments on. For thee algorithms that we are able to do so, we use our baseline 3 layer neural network as the model architecture.

\smallsec{RWT~\cite{jiang2020reweight}} reweighting scheme on the training samples that learns group-specific weights between each group's positive and negative instances. The algorithm lowers the weight on positive examples of a group if its TPR is higher than the overall rate, and increases the weight on the positive examples otherwise. In the original algorithm this is an iterative process whereby the entire classifier is retrained with each new set of weights. In extending this method to a neural network, 
we continue training the model at each iteration \emph{without} retraining the whole model from scratch.

\smallsec{RDC~\cite{agarwal2018reductions}} reduces optimizing for both accuracy and a fairness constraint to a sequence of cost-sensitive classifications, which can be solved to yield a randomized classifier. We adapted this to yield continuous outputs, by using the probabilities output by each classifier in the ensemble.

\smallsec{LOS~\cite{foulds2018intersectionality}} weighted addition to loss of an extra intersectional fairness regularizing term that minimizes the maximum log ratio between the rate of positive classification of all groups. In order to modify this for our fairness criterion, we minimize the maximum ratio between the TPR of all groups.

\smallsec{GRP~\cite{yang2020overlap}} GroupFair is probabilistic combinations of logistic regression models that ensure fairness for overlapping groups. This method contains two variations, weighted ERM and plugin, and we only use the latter due to the prohibitively long computational time required for the former.
We combine the model's continuous outputs rather than discrete ones. 

\smallsec{GRY~\cite{kearns2018gerry, kearns2018empirical}} cost-sensitive classifications to obtain solutions to a 2-player zero-sum game between a \emph{learner} (which learns the classifier) and an \emph{auditor} (which ensures that the fairness criterion is met). The method produces a sequence of classifiers, all of which output hard outputs, and we use the average of these outputs. We use linear regression for the individual models. 
\\

Hyperparameters are all tuned on the validation set. The first split is 70-30 for training/validation and the test set. The training/validation is further split at 70-30 again to make the training and validation sets.

\smallsec{Baseline} batch size: 64, and hyperparameter tuning across epochs: [50, 100, 150] $\times$ learning rate: [.001, .005]

\smallsec{RWT~\cite{jiang2020reweight}} batch size: 64, learning rate: .001, hyperparameter tuning across epochs: [100, 150] $\times$ reweight learning rate (algorithm-specific hyperparameter): [.1, .2, .5, 1.] 

\smallsec{RDC~\cite{agarwal2018reductions}} hyperparameter tuning across batch size: [256, 512, 1024, 2048] $\times$ epochs: $[50, 100, 200] \times$ number of iterations:[10, 20, 50]

\smallsec{LOS~\cite{foulds2018intersectionality}} batch size: 1024, learning rate: .005, hyperparameter tuning across epochs: [200, 250, 300] $\times$ $\lambda$ weight on additional loss: [.01, .5, .1] 

\smallsec{GRP~\cite{yang2020overlap}} 
epochs: 10000, learning rate: .01, B: 50, hyperparameter tuning across $\nu$: [0.001, 0.003, 0.01, 0.03, 0.1]

\smallsec{GRY~\cite{kearns2018gerry, kearns2018empirical}} hyperparameter tuning across C:[5, 10, 20] $\times$ number of iterations: $[50, 100, 200] \times$ fairness parameter $\gamma$:[1e-3, 5e-3, 1e-2]

\end{document}